\documentclass{article}

\usepackage[english]{babel}

\usepackage[letterpaper,top=2cm,bottom=2cm,left=3cm,right=3cm,marginparwidth=1.75cm]{geometry}

\usepackage{amsmath}
\usepackage{graphicx}
\usepackage{booktabs}
\usepackage{amsmath,amssymb}
\usepackage{amssymb}
\usepackage{latexsym}
\usepackage {amsmath}
\usepackage{xcolor}
\usepackage{hyperref}
\usepackage{multirow}
\usepackage{multicol}
\usepackage{array}
\usepackage{authblk}
\usepackage[numbers,sort&compress]{natbib}

\title{A Cognitive Explainer for Fetal ultrasound images classifier Based on Medical Concepts}

\author[1]{Yingni Wang}
\author[1]{Yunxiao Liu}
\author[2]{Licong Dong}

\author[1]{Xuzhou Wu}
\author[3]{Huabin Zhang}
\author[4]{Qiongyu Ye}
\author[2]{Desheng Sun}
\author[5]{Xiaobo Zhou}
\author[1]{Kehong Yuan}

\affil[1]{Tsinghua Shenzhen International Graduate School, University Town of Shenzhen, Nanshan District, Shenzhen and 518055, China}
\affil[2]{Department of Ultrasound, Beijing Tsinghua Changgung Hospital, Beijing and 102218, China}
\affil[3]{Department of Ultrasonography, Peking University Shenzhen Hospital, Shenzhen and 518055, China}
\affil[4]{Department of Ultrasonography, Shenzhen Baoan District Maternal and Child Health Hospital, Shenzhen and 518055, China}
\affil[5]{Center for Computinal Systems Medicine of SBMI, UTHealth, Houston,TX 75835,USA}

\date{} 

\begin{document}
\maketitle

\begin{abstract}
Fetal standard scan plane detection during 2-D mid-pregnancy examinations is a highly complex task, which requires extensive medical knowledge and years of training. Although deep neural networks (DNN) can assist inexperienced operators in these tasks, their lack of transparency and interpretability limit their application. Despite some researchers have been committed to visualizing the decision process of DNN, most of them only focus on the pixel-level features and do not take into account the medical prior knowledge. In this work, we propose an interpretable framework based on key medical concepts, which provides explanations from the perspective of clinicians' cognition. Moreover, we utilize a concept-based graph convolutional neural(GCN) network to construct the relationships between key medical concepts. Extensive experimental analysis on a private dataset has shown that the proposed method provides easy-to-understand insights about reasoning results for clinicians. 
\end{abstract}


\section{Introduction}
\label{sec1}
Ultrasonography is widely used for the prenatal assessment of growth and anatomy, which can provide diagnostic findings that often contribute to the management of problems in later pregnancy\cite{salomon2011practice}. Due to the low cost, wide availability, and non-invasiveness, the 2D ultrasound (US) is the primary modality for the evaluation of the fetus's health\cite{baumgartner2017sononet}. Currently, to improve the quality of the population, most countries offer at least one mid-trimester scan\cite{salomon2011practice}. 

Manipulating the probe to obtain the standard scan plane in variable anatomy and assessing the hard-to-understand US data are highly sophisticated tasks, requiring years of training\cite{maraci2014searching}. 
Moreover, the diagnostic accuracy of obstetric sonography is related to the inherent limitations of US technology, which is operator-dependent and short of consistency, standardization, and reproducibility. In some cases, it can be difficult to obtain a desired standard plane if the fetal pose is inapposite\cite{abuhamad2008automated}. It is even a challenging task for inexperienced operators and non-experts to identify the relevant structures in a given standard plane for certain views. Furthermore, there is a striking shortage of experienced operators, with vacancy rates reported to be as high as $18.1 \%$ in the UK\cite{Sonographer}.

The rapid development of deep convolution neural networks (CNNs) has achieved great performance in medical image diagnosis of many diseases such as lung nodule\cite{huang2017lung,zuo2019multi,su2021lung}, pancreatic cancer\cite{sekaran2020deep,liu2020deep,li2021detection} and Alzheimer\cite{khvostikov20183d,farooq2017deep,salehi2020cnn}. While the automatic detection of fetal standard scan planes has been explored\cite{baumgartner2017sononet,ryou2016automated,chen2015automatic,li2018standard,chen2015standard},most of these methods simply employ black-box models to directly obtain  correct standard scan planes. Despite their state-of-art performance, the lack of interpretability and transparency greatly hinder their clinical application. On the other hand, medical decisions may have life-or-death consequences, medical diagnosis applications require not only high performance but also a strong rationale for judgment\cite{kim2021xprotonet,zhou2022ted,mahapatra2022interpretability}. 

In recent years, the interpretability technique for CNNs has emerged as an important research topic with substantial progress. And most explainable approaches attempt to explain CNN with saliency \cite{zhou2016learning,selvaraju2017grad,rebuffi2020there,jalwana2021cameras,lee2021relevance,simonyan2013deep}, perturbation-base \cite{petsiuk2021black,shi2019explainable,sundararajan2017axiomatic}, and logical-based \cite{alaniz2021learning,nauta2021neural} methods. There are still some problems with these algorithms in their application: (1) They only provide pixel-level explanations between input and output, and do not take into account the relationship between anatomical structures; (2) pixel-level explanations tend to be blurry and hard-to-identify for radiologists; (3) These methods do not provide a systematic assessment of the explanations.

With this in mind, we propose a cognitive explainer for fetal US scan planes based on medical concepts, enabling CNN to explain from the perspective of the sonographer’s cognition. We first select three important standard planes as our research objects, namely the Fetal Abdominal Standard plane(FASP), Fetal Thalamus Standard Plane(FTSP), and Fetal Femur Standard plane(FFSP). The first stage of our framework aims to extract medical concepts with a simple linear iterative clustering algorithm. Then we employ the anatomical prior knowledge provided by the sonographer, including position, shape, texture, brightness, etc., to locate the key medical concepts. After that,  a GCN is utilized to model the interaction between these concepts, simulating the decision-making process of doctors. 

\begin{figure}[t]
\centering
\includegraphics[width=0.8\columnwidth]{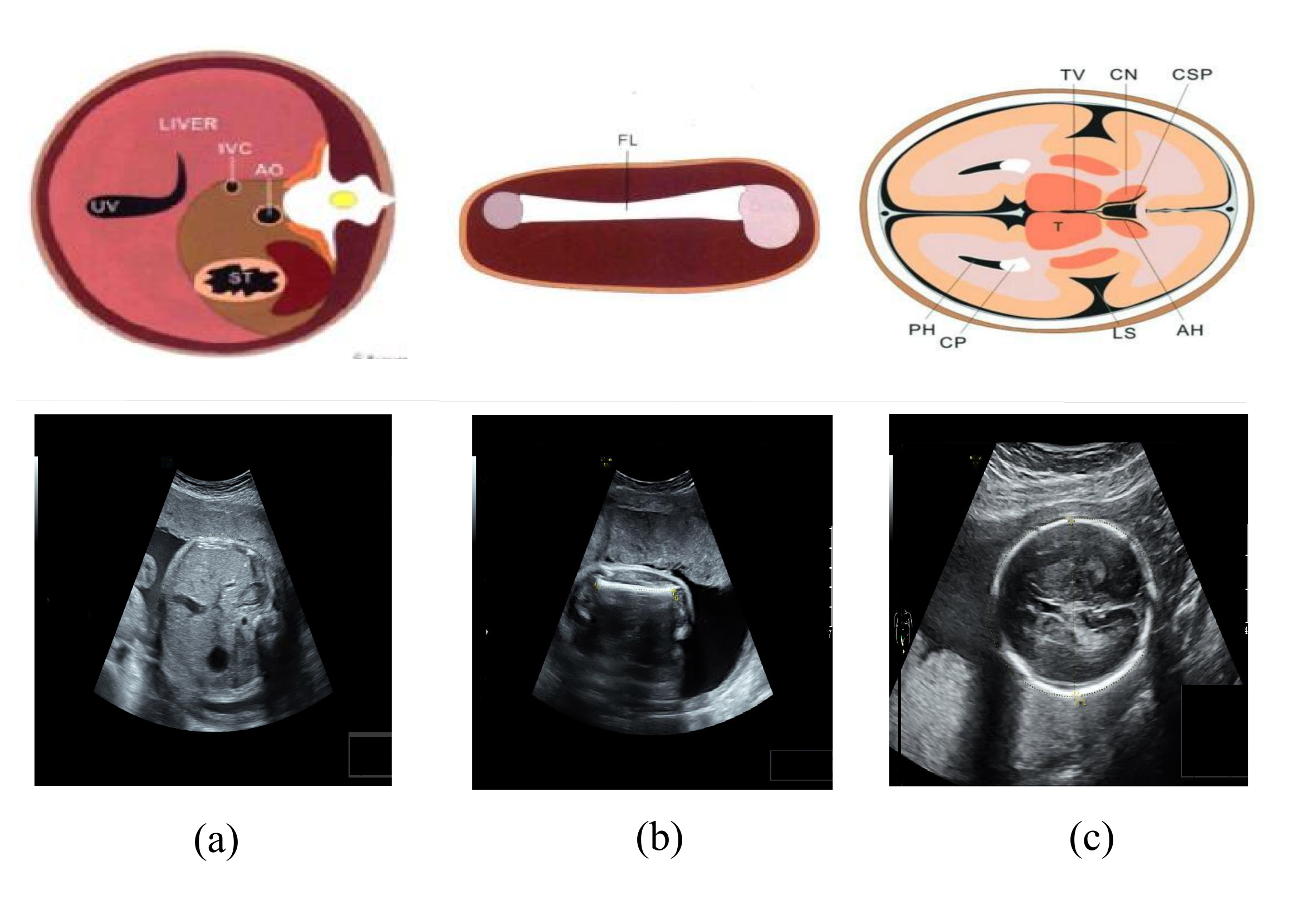} 
\caption{Fetal abdominal, femoral, and thalamic anatomy (the top row) and typical standard scan planes(the bottom row). }
\label{fig1: the examples}

\end{figure}

In summary, the specific contributions of our work are:
\begin{enumerate}
\item We propose an interpretable framework for fetal US standard plane classification based on medical concepts, which are identified by medical prior knowledge.
\item We use medical concepts that sonographers care about and their relative relationships to construct GCN, encode the spatial positions between them, and provide interpretation from the perspective of sonographers' cognition.
\item Extensive qualitative and quantitative assessment of various graph explainability techniques in US, with a validation of the findings by expert sonographers.

\end{enumerate}

\section{Related Work}

\subsection{Medical concepts for fetal standard scan plane diagnosis}
The sonographic parameters in FASP, FTSP, FFSP can be used to estimate gestational age and for fetal size assessment\cite{degani2001fetal,altman1997new}.Correct US standard plane scanning is the basis for precise measurement of clinical parameters. 
Clinically, to locate the FASP, a sonographer attempts to find the concurrent presence of the key anatomical structures: the stomach bubble (SB), the umbilical vein (UV) and the spine (SP) when moving the transducer across the pregant woman's body\cite{chen2015standard}. Similarly, the key anatomical structures in FTSP are the cavity of the septum pellucidum (CSP), the right thalamus(RT), and the left thalamus(LT). And the key anatomical structures in FFSP are the femur(FM) and mataphysis(MP). The anatomy structure and typical standard planes are illustrated in Fig~\ref{fig1: the examples}.

\subsection{Machine-aided diagnosis of fetal US planes}
Recently, several machine learning methods have been proposed to address the US standard plane classification tasks\cite{maraci2015fisher,maraci2014searching}. The earlier works mostly rely on extracting hand-crafted features or incorporated component-based geometric constraints\cite{kwitt2013localizing,rahmatullah2012integration,ni2014standard}. Motivated by the development of computer vision, CNNs are more employed in the analysis of US data. \citet{chen2015automatic} presented a framework to detect standard planes from US videos automatically, which explores spatiotemporal features learning with a novel knowledge-transferred recurrent neural network. After that, SonoNet\cite{baumgartner2017sononet} only based on image-level labels can not only automatically detect 13 fetal standards views in 2-D US images, but also provide localization of the fetal structures via a bounding box. \citet{ryou2016automated} presented a framework to localize the fetus and extract the fetal biometry planes for the head and abdomen in 3D fetal US by breaking down the 3D volume into a stack of 2D slices. Then transfer learning CNNs were applied to classify standard planes. Different from the methods proposed above, iterative Transformation Network  \cite{li2018standard} approached the standard plane detection problem by regressing rigid transformation parameters. And they used a CNN to learn the relationship between a 2D plane image and the transformation parameters required to move that plane towards the location/orientation of the standard plane in the 3D volume.

\subsection{Interpretability methods in medical application}

During the past few years, the application of DNNs for automatic diagnosis of medical diseases has shown a good prospect. At the same time, many works have been devoted to improving the interpretability and transparency of neural networks. At present, the interpretability methods can be divided into two categories: (1)explaining a neural network posthoc; and (2) building an inherently interpretable model \cite{donnelly2022deformable}. Examples of the posthoc methods include  activation-based\cite{das2020interpreting,kim2018interpretability}, perturbation-based\cite{dabkowski2017real,fong2017interpretable,petsiuk2018rise,ribeiro2016should,elliott2021explaining}, and backpropagation-based approaches\cite{zhou2016learning,selvaraju2017grad,rebuffi2020there,lee2021relevance,jalwana2021cameras}. And the inherently interpretable models include prototype-based networks\cite{nauta2021neural,donnelly2022deformable}, BagNets\cite{brendel2019approximating}, CoDANets\cite{bohle2021convolutional} and B-cos\cite{bohle2022b}.

CheXNet\cite{rajpurkar2017chexnet} localized pathologies it identified using CAM, which highlights the areas of the X-ray that are most important for making a particular pathology classification. Based on the assumption that thorax disease usually happens in localized areas and the existence of irregular borders hinders the network performance, a three-branch attention-guided convolution neural network (AG-CNN)\cite{guan2018diagnose} integrates a global branch to compensate the lost discriminative cues by the local branch. \citet{tang2019interpretable} attempted to use Guided Grad-CAM and feature occlusion to visualize the feature salience in identifying specific neuropathologies-amyloid plaques and cerebral amyloid angiopathy-in immunohistochemically-stained archival slides.
\citet{khakzar2021towards} use Network Dissection to quantify the interpretability of chest X-ray classification models. \citet{schuchmacher2021framework} introduces a hypothesis-based framework for falsifiable explanations of machine learning models, which connects an intermediate space induced by the model with the inputs.

Furthermore, some methods master the ability to automate the human-like diagnostic reasoning process and translate gigapixels directly to a series of interpretable predictions, providing second opinions and encouraging consensus in clinics\cite{zhang2019pathologist}. \citet{jaume2021quantifying} adopted a biological entity-base graph and yielded intuitive pathological interpretability. They also proposed a set of novel quantitative metrics based on statistics of class separability using pathologically measurable concepts to characterize graph explainers, which relaxes the exhaustive assessment by expert pathologists. Several studies attempt to learn representative features of the disease and make decisions based on these features. Interpretable CNN models are designed to operate in a human-understandable manner \cite{rudin2019stop}. XProtoNet \cite{kim2021xprotonet} learns representative patterns of each disease from X-ray images and makes a diagnosis on a given X-ray image based on patterns.  \citet{zhou2022ted} propose a Two-stage Expert-guided Diagnosis framework to simulate the radiologists' decision process. They utilize the key imaging attributes in the first stage as a form of attention and soft supervision through a variant of triplet loss. During the training process, the network learns more semantically correlated representations and increases its interpretability. \citet{mahapatra2022interpretability} propose an interpretability-guided inductive bias approach enforcing the learned features yield more distinctive and spatially consistent saliency maps. These works targeted classification tasks in X-ray and pathological images, and there was no attempt to make an interpretable automated diagnosis framework for fetal US standard plane identification. To this end, we propose an interpretable classification model for the fetal US standard plane that learns the important structure between medical concepts and location relationship .

However, most of these interpretability methods are based on pixel-wise relations, which are different from human cognition. Several recent studies have attempted to extend the interpretability methods to visual concepts that humans intuitively understand. \citet{zhang2019interpreting} presents an explicit visual reasoning method, which incorporates external knowledge and models high-order relational attention. After that, \citet{ge2021peek} proposed a visual reasoning explanation framework based on structural concept graphs to answer interpretability questions and potentially provide guidance on improving DNN’s performance.  Although some interpretation methods based on visual concepts, such as ACE\cite{ghorbani2019towards} and TACV \cite{kim2018interpretability}, are very effective in extracting natural images, their performance in medical images is poor. Our methods also require extracting visual concepts, but it is more adaptable to medical tasks. By combining medical prior knowledge, the meaning of concepts can be more clearly defined and the explanation obtained is more consistent with clinical experience.

\subsection{Graph Neural Networks}
GCN\cite{chiang2019cluster,schlichtkrull2018modeling,yue2020graph} can process non-Euclidean structured data and model complex information in the graphs, such as heterogeneous connections and high-order connections. In recent years, graph neural networks are often used to learn high-order relationships. For example, \citet{zhao2021graph}proposed an effective graph-based relation discovery approach to build a contextual understanding of high-order relationships. On the other hand, thanks to its powerful information processing capabilities, Graph neural networks (GNNs) have been widely applied in many visual and linguistic tasks, such as VQA \cite{ghosh2019generating,teney2017graph,shi2019explainable}, image captioning \cite{gu2019unpaired,xu2019scene,yang2019auto}, and scene understanding tasks\cite{li2017scene}. In this study, we use GCN to capture high-order semantic relationships between key medical concepts and provide more credible explanations for clinical diagnosis.

\section{Method}

\begin{figure*}[ht]
\centering
\includegraphics[width=0.78\textwidth]{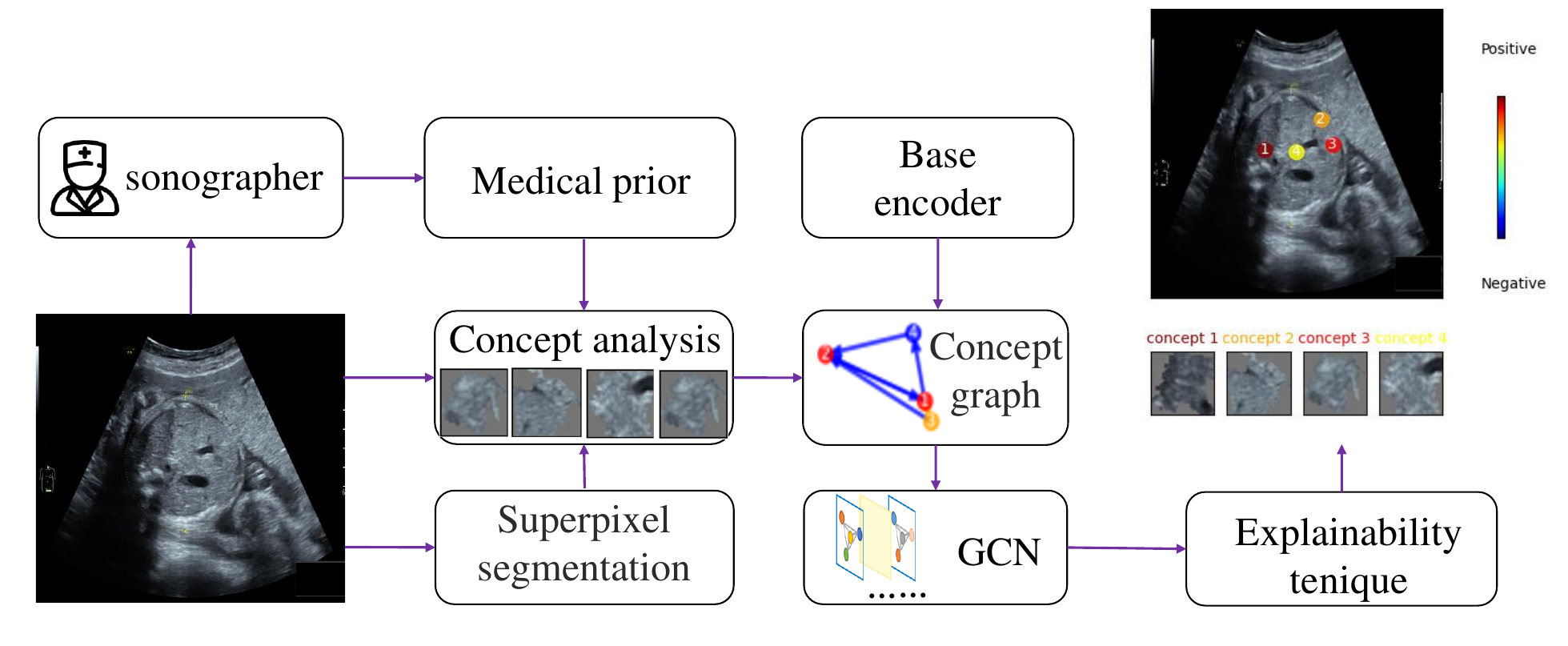} 
\caption{Overview of the interpretability framework. We first extract class-specific medical concepts approved by sonographers with prior medical knowledge. The medical concepts are transformed into graph-structured data, and use GCN to learn the contribution of nodes (medical concepts) and edges (relationships between concepts) to decision-making and to explain the decision-making process of the network.
}
\label{fig2:Overview of our refinement framework}
\end{figure*}

The overall pipeline of the proposed framework is presented in Fig.~\ref{fig2:Overview of our refinement framework}. First, we extract medical concepts approved by sonographers, combined with medical prior knowledge. Then, we transform these concepts into graph-structured data. Next, we introduce a "black-box" GCN that maps the graph to the corresponding class label. Finally, we utilize a post-hoc explanation technique to interpret the decision mechanism of the network and visualize the reasoning process.

\subsection{Medical concepts identification with prior knowledge}
When performing a US scan, instead of directly identifying the standard plane,  sonographers often first searches for necessary evidence to support their decision. The first step of our pipeline attempts to mimic the sonographer's reasoning process and discover key anatomical structures for standard plane identification.  First, we employ a simple linear iterative clustering algorithm (SLIC) to obtain the candidates for the medical concepts. Then we adopt a divide-and-conquer strategy and develop a selection scheme based on the characteristics of the different standard planes.

Specifically, the anatomical structures in FASP and FTSP have more complex morphology and are difficult to identify. Therefore, we first utilize a pretrained segmentation model to obtain the abdominal and head circumference, whose shape is close to ellipses. Based on the anatomy graph in Fig~\ref{fig1: the examples}, we discover that the UV and SP are approximately located near the major axis of the ellipse, and the SB located near the minor axis of the ellipse. Moreover, the SB is generally oval and has low brightness. The SP often appears as a coarser texture in the image. The US also has a low pixel value in the image. Based on that knowledge, we can locate medical concepts that are important discriminatively for classification.  As for LT and RT, they are also located around the major axis of the ellipse. 

The key step in converting an US image to graph-structured data is to extract medical concepts from the US image that are important for differentiation and diagnosis. This ensures that the inputs to our method are medically interpretable and can be directly linked and reasoned with by the sonographer. Inspired by VRX\cite{ge2021peek} and XProtoNet \cite{kim2021xprotonet}, we use visual concepts to represent an input image given class-specific knowledge of the pretrained combined medical prior knowledge with visual concept extraction to find the representative features. 
First, multi-resolution segmentation methods were applied to extract the superpixels containing the medical concepts. While ACE is reasonably effective in extracting visual concepts in natural images, its performance is poor on US images. Due to the small foreground area of medical images, most superpixels obtained by multi-resolution methods are backgrounds and irrelevant tissues, which increases the difficulty of concept extraction. To alleviate this issue, given an image $I$, we use Grad-CAM to generate attention heatmaps and constrain the target area in the foreground, thereby helping us exclude irrelevant areas for diagnosis.

\subsection{Concept graph construction}
We define a medical concept graph $G:=(V,E)$  as a set of nodes $V$ and edges $E$ . $v_{i} \in V$ denotes a relevant medical concept. The node attributes are the high-order feature encoded by a trained CNN 
$F( \cdot )$ classifier  . The properties of directed edges $e_{ij}=(v_{i},v_{j})$  in the graph have two meanings: 1) The relative relationships of each concept in space, which is initialized by the relative position of two concepts in the image. 2) The correlation between two concepts which ares is given by medical prior knowledge. This is essential to the reasoning process of medical concepts.

Based on the above physician-approved medical concepts, we use a trained CNN classifier $F(\cdot)$ to extract the high-order features of medically separable visual concepts and take them as attributes of graph nodes.
Given an input image $I$, $X \in R^{C \times H \times W}$ are the feature maps encoded by a trained CNN classifier $F(\cdot)$.

\subsection{Concept graph Learning}
\label{GCN}
Given $G$ , graph-structured data of US images, we aim to infer the corresponding class of standard section. 
We use GraphConv\cite{morris2019weisfeiler} as the backbone of our network. A layer from GCN has two steps: message aggregation and update. Formally, we define a layer as: 
\begin{equation}
    x^{i}_{k+1}=W_{1}h^{i}_{k}+\sum_{i \in N(i)}W_{3}C(\alpha^{c}_{ji}W_{2}x^{j}_{k},e^{ji}_{k})
\end{equation}

\begin{equation}
    e^{ji}_{k+1}=W_{4}x^{ji}_{k}
\end{equation}
where $x^{i}_{k}$  denotes the features of a node $v_{i}$  in layer k,   $W_{1}$ and $W_{2}$ denotes the shared transformation parameters for the target node  $v_{i}$ and its neighbor node $v_{j}$  respectively. $W_{3}$  and $W_{4}$  denote a linear transformation for edge features, $N_{i}$  denotes the neighboring nodes of $v_{i}$ . $C(\cdot)$  denotes concatenation. $\alpha^{c}_{ji}$ is a predefined prior coefficient, which measures the interdependence between various concepts. 
After N iterations, we use an MLP to process graph features and generate an n-dimensional probability distribution vector.

\subsection{Post-hoc graph explainer}
As shown in Fig.~\ref{fig2:Overview of our refinement framework}, after an image is fed into our GCN, we can obtain the prediction score $\hat{y}=\Phi(G(x))$ . Where $\Phi$  is graph embedding networks and with $m$ layers. We generate the explanation per concept graph by employing post-hoc graph explainers. We can evaluate the anatomical relevance of the black-box neural network reasoning based on the explanations. In this work, we consider three types of graph explainers for explaining concept graphs, which follow similar operational settings, i.e. (i) input data are concept graphs, (ii) a GCN is trained a priori to classify the input data. 
\begin{enumerate}
\item Graph Sensitivity Analysis(SA)

SA\cite{baldassarre2019explainability} is a backpropagation-based saliency method, which produces the explanation of a black-box model using the squared norm of its gradient w.r.t. the inputs $x$. Inspired by SA, for a specific class $c$,we obtain its corresponding prediction score $\hat{y}_{c}$  and calculate the gradients concerning the graph of each layer in GCN as: 
\begin{equation}
    w_{i}=\frac{\partial \hat{y}_{c}}{\partial G_{i}(x_{i})}
\end{equation}
The contribution score of each node(medical concept) or edge to the network’s decision is computed as follows:
\begin{equation}
    s_{i}=w^{T}_{i}G(x_{i})
\end{equation}

\item Graph Integrated Gradients(IG)

To solve the problem of breaking sensitivity of the gradients, 
 IG\cite{sundararajan2017axiomatic}
obtain the importance of features in a black-box model by examining the gradients of the counterfactuals obtained by scaling the input. Similarly, the importance score of concept $x_{i}$ in GCN can be computed as follows. Here, $x_{i}^{'}$ is the baseline concept inputs, and $\frac{\partial G(x)}{\partial x_{i}}$ is the gradient of $G(x)$. 

\begin{equation}
    s(x_{i})::=(x_{i}-x_{i}^{'}) \times \int_{\alpha =0}^{1} \frac{\partial G(x_{i}^{'}+\alpha \times (x_{i}-x_{i}^{'}))}{\partial x_{i}}
\end{equation}

\item Graph Grad-CAM

Different from the first interpretability methods, Grad-CAM\cite{selvaraju2017grad} explored visualizing saliency at intermediate layers by combining information from activations and gradients. It produces class activation maps based on these two steps. First, it 
\begin{equation}
    w_{i}=\frac{1}{Z} \sum_{i}\sum_{j} \frac{\partial \hat{y}_{c}}{\partial G(x_{i,j})}
\end{equation}

Similarly, the contribution score in the concept space according to Grad-CAM is computed as:
\begin{equation}
    s_{i}=ReLU(\sum_{k} w_{i} x_{i,j})
\end{equation}

\end{enumerate}

\section{Datasets and Implementation details}

\subsection{Datasets}
Our datasets were all from the Department of US, Shenzhen Hospital, Peking University (Hospital A), and Shenzhen Baoan District Maternal and Child Health Hospital(Hospital B). We collected retrospectively two-dimensional  FFSP, FTHP, and FASP from 1070 examined pregnant women with a total of 938, 1201 and 1337 US iamges respectively. In addition, our dataset also contained 1988 of 'other' views. US images were acquired using GE Voluson E6, E8, E10 color Doppler US diagnostic system, Sonospace, and Mindray Resona R9S. For each scan, we had access to freeze-frame images saved by the sonographers during the exam. The minimum gestational age of the fetus is 16 weeks and the maximum is 39 weeks. Details are shown in Table~\ref{table1:The details about our datasets.}. 

\begin{table}[t]
\centering
\begin{tabular}{m{1.8cm}<{\centering} m{1.2cm}<{\centering} m{1.2cm}<{\centering} m{1.2cm}<{\centering} m{1.2cm}<{\centering}}
\hline
Dataset & FASP & FFSP & FTSP & other   \\
\hline
Hospital A & 1273 & 887 &1138 & 1988\\
Hospital B & 64 & 51 & 63 & 52\\
\hline
\end{tabular}
\caption{The details of experimental datasets.}
\label{table1:The details about our datasets.}
\end{table}

\subsection{Implementation details}

\begin{figure}[ht]
\centering
\includegraphics[width=0.5\textwidth]{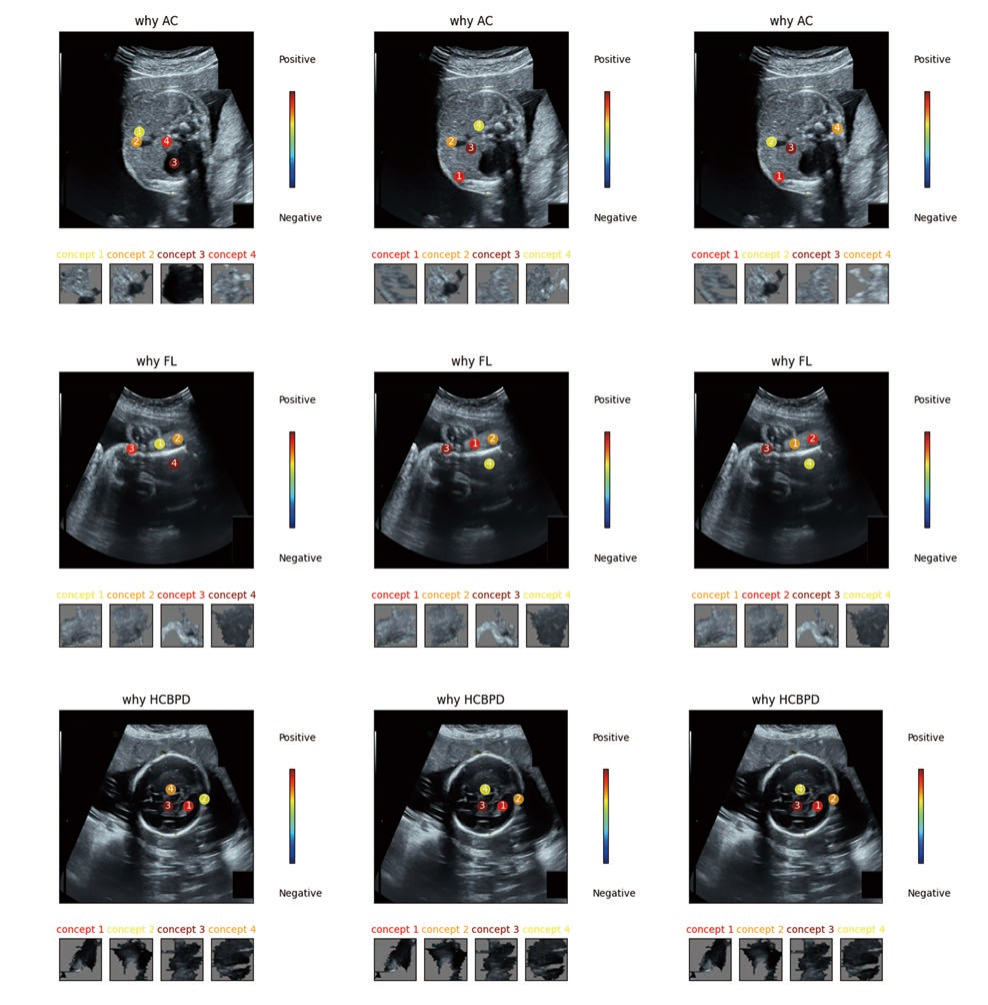} 
\caption{Medical concept reasoning explanation of Fetal US Standard Plane for MobilenetV2 in B dataset. The rows represent the FSP types, i.e. FASP, FFSP, and FTSP, and the columns represent the graph interpretability methods, i.e. Graph SA, Graph IG and Graph Grad-CAM. Concept importance ranges from blue (the least important) to red (the most important).
}
\label{fig3:Medical concept reasoning explanation}
\end{figure}

We conducted our experiments using PyTorch in 11GB NVIDIA GeForce RTX 2080Ti GPU. We conducted our experiments using Pytorch and the PyTorch Geometric Library. The GCN architecture is presented in Section ~\ref{GCN}. The CNN classifier was trained for 100 epochs by Adam optimizer\cite{adam}, $10^{-2}$ learning rate, 32 batch size, and dynamic adjustment. As for GCN classifiers, they were trained for 200 epochs by Adam optimizer\cite{adam}, $10^{-2}$ learning rate, 128 batch size, and dynamic adjustment. To prevent our model from learning the manual annotations located in the US images by the sonographers rather than the images themselves, we remove all the annotations, vendor logos, and US control indicators based on HSV space and location prior. Furthermore, we normalize each image by subtracting the mean intensity value and dividing it by the image pixel standard deviation.

To make the most use of our data while evaluating the generalization performance, We divided the data from hospital A into a training set, validation set, and test set at a ratio of 7:2:1. And the data from Hospital B were not involved in the training and were used to evaluate the generalization ability of the model.

\subsection{Evaluation metrics}
Considering the imbalance class in the dataset, we utilize the following metrics to evaluate the performance of CNN and GCN networks, i.e. Accuracy (ACC), Precision, Recall, and F1 score. 

\begin{equation}
    ACC=\frac{TP+TN}{TP+TN+FP+FN}
\end{equation} 

\begin{equation}
    Precision = TP/(TP+FP)
\end{equation} 

\begin{equation}
    Recall = TP/(TP+FN)
\end{equation} 

\begin{equation}
    F1 = \frac{2*Precision*Recall}{Precision+Recall}
\end{equation} 

Here $TP$, $FN$, $TN$, and $FP$ denote the number of true positive, false negative, true negative, and false positive respectively. And the area under the curve (AUC) is also adopted.

\section{Experiments And Results}
This section describes the interpretability analysis of GCN for the classification of the fetal standard US plane. In our experiments, we use VGG\cite{zeiler2014visualizing},  ResNet\cite{he2016deep}, mobilenetV2\cite{DBLP:journals/corr/abs-1801-04381}, and DenseNet{\cite{densenet}} as the basic neural networks.

\subsection{Quantitative results}
In order to quantitatively assess the classification performance of the CNN and GCN models, we evaluated these networks on the test datasets and datasets from Hospital B.  In table ~\ref{table2:Accuracies across validation datasets of CNN} and ~\ref{table3:Accuracies across baoan datasets of CNN} we report the average scores for all examined CNN networks in test dataset and dataset from Hospital B. Similarly, the average scores for all examined GCN  networks in test dataset and dataset from Hospital B were reported in Table ~\ref{table4:Accuracies across validation datasets of GCN} and Table ~\ref{table5:Accuracies across baoan datasets of GCN}. 

From table ~\ref{table2:Accuracies across validation datasets of CNN} and ~\ref{table3:Accuracies across baoan datasets of CNN} it can be seen that almost all networks achieved consistent performance on the test dataset, with minor differences for VGG19. However, for the dataset from Hospital B, ResNet50 has outstanding performance in all five metrics. 

\begin{figure}[ht]
\centering
\includegraphics[width=0.4\textwidth]{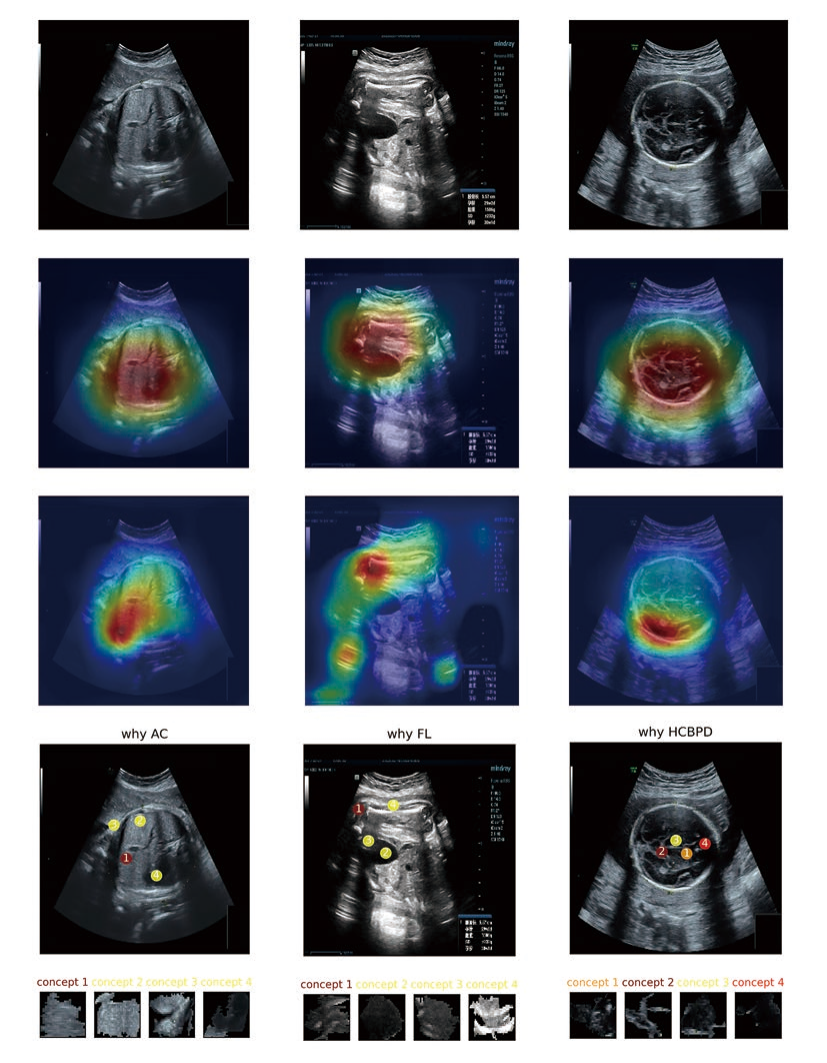} 
\caption{ Comparison of explainability methods on Fetal Standard Planes based on MobilenetV2.
}
\label{fig5: Comparison of MobilenetV2}
\end{figure}

On the other hand, as shown in Table ~\ref{table4:Accuracies across validation datasets of GCN} and Table ~\ref{table5:Accuracies across baoan datasets of GCN}, for GCN classifier, Densenet121, and MobilenetV2  performed very similarly on the test dataset with Densenet121 obtaining slightly better ACC, precision and recall score as well as AUC. As for B dataset, ResNet34 and MobilenetV2 obtained very closed classification scores but ResNet34 performed poorly in ACC, AUC, and F1.

\begin{table*}[ht]
\centering
\begin{tabular}{m{2cm}<{\centering} m{1.4cm}<{\centering} m{1.4cm}<{\centering} m{1.4cm}<{\centering}  m{1.4cm}<{\centering} m{1.4cm}<{\centering}}
\hline
model & ACC & precision & recall & AUC & F1 \\
\hline
ResNet18 & 1.0 $\%$ & 1.0 $\%$ & 1.0 $\%$ & 1.0 $\%$ & 1.0 $\%$\\
ResNet34 & 1.0 $\%$ & 1.0 $\%$ & 1.0 $\%$ & 1.0 $\%$ & 1.0 $\%$\\
ResNet50 & 1.0 $\%$ & 1.0 $\%$& 1.0 $\%$ & 1.0 $\%$ & 1.0 $\%$\\
VGG16 & 1.0 $\%$ & 1.0 $\%$& 1.0 $\%$ & 1.0 $\%$ & 1.0 $\%$\\
VGG19 & 99.43 $\%$ & 99.31 $\%$& 99.33 $\%$ & 1.0 $\%$ & 99.32 $\%$\\
MobilenetV2 & 1.0 $\%$ & 1.0 $\%$& 1.0 $\%$ & 1.0 $\%$ & 1.0 $\%$\\
Densenet121 & 1.0 $\%$ & 1.0 $\%$& 1.0 $\%$ & 1.0 $\%$ & 1.0 $\%$\\
\hline
\end{tabular}
\caption{Classification scores for the CNN classification models in datasets from Hospital B. }
\label{table2:Accuracies across validation datasets of CNN}
\end{table*}

\begin{table*}[ht]
\centering
\begin{tabular}{m{2cm}<{\centering} m{1.4cm}<{\centering} m{1.4cm}<{\centering} m{1.4cm}<{\centering}  m{1.4cm}<{\centering} m{1.4cm}<{\centering}}
\hline
model & ACC & precision & recall & AUC & F1 \\
\hline
ResNet18 & 86.52 $\%$ & 86.90 $\%$ & 85.37 $\%$ & 96.89 $\%$ & 84.09 $\%$\\
ResNet34 & 83.48 $\%$ & 83.58 $\%$ & 83.07 $\%$ & 95.86 $\%$ & 83.19 $\%$\\
ResNet50 & \textbf{90.00} $\%$ & \textbf{89.36} $\%$& \textbf{89.55} $\%$ & \textbf{97.71} $\%$ & \textbf{89.28} $\%$\\
VGG16 & 86.52 $\%$ & 86.90 $\%$ & 84.41 $\%$ & 93.72 $\%$ & 82.89 $\%$\\
VGG19 & 85.65 $\%$ & 86.12 $\%$ & 84.40 $\%$ & 93.72 $\%$ & 82.89 $\%$\\
MobilenetV2 & 85.22 $\%$ & 84.42 $\%$ & 84.67 $\%$ & 97.45 $\%$ & 84.46 $\%$\\
Densenet121 & 87.83 $\%$ & 88.01 $\%$ & 86.81 $\%$ & 97.22 $\%$ & 85.79 $\%$\\
\hline
\end{tabular}
\caption{Classification scores for the CNN classification models in validition datasets. Results are taken from our trained model.}
\label{table3:Accuracies across baoan datasets of CNN}
\end{table*}

\begin{table*}[ht]
\centering
\begin{tabular}{m{2cm}<{\centering} m{1.4cm}<{\centering} m{1.4cm}<{\centering} m{1.4cm}<{\centering}  m{1.4cm}<{\centering} m{1.4cm}<{\centering}}
\hline
model & ACC & precision & recall & AUC & F1 \\
\hline
ResNet18 & 94.06 $\%$ & 93.05 $\%$ & 92.74 $\%$ & 92.86 $\%$ & 99.40 $\%$\\
ResNet34 & 93.68 $\%$ & 92.59 $\%$ & 92.41 $\%$ & 92.38 $\%$ & 99.36 $\%$\\
ResNet50 & 92.91 $\%$ & 91.92 $\%$& 91.70 $\%$ & 91.80 $\%$ & 99.50 $\%$\\
VGG16 & 86.52 $\%$ & 84.27 $\%$ & 83.75 $\%$ & 83.60 $\%$ & 97.48 $\%$\\
VGG19 & 88.46 $\%$ & 87.21 $\%$ &  87.40 $\%$ & 87.17 $\%$ & 98.12 $\%$ \\
MobilenetV2 & 96.17 $\%$ & 95.81 $\%$ & 95.56 $\%$ & 95.68 $\%$ & \textbf{99.81} $\%$\\
Densenet121 & \textbf{96.74} $\%$ & \textbf{96.12} $\%$ & \textbf{96.26} $\%$ & \textbf{96.19} $\%$ & 99.80 $\%$\\
\hline
\end{tabular}
\caption{Classification scores across all GCN classification models in test datasets. }
\label{table4:Accuracies across validation datasets of GCN}
\end{table*}

\begin{table*}[ht]
\centering
\begin{tabular}{m{2cm}<{\centering} m{1.2cm}<{\centering} m{1.2cm}<{\centering} m{1.2cm}<{\centering}  m{1.2cm}<{\centering} m{1.2cm}<{\centering}}
\hline
model & ACC & precision & recall & AUC & F1 \\
\hline
ResNet18 & 81.82 $\%$ & 84.45 $\%$ & 81.53 $\%$ & 81.59 $\%$ & 93.36 $\%$\\
ResNet34 & \textbf{85.39} $\%$ & 85.51 $\%$ & \textbf{85.24} $\%$ & 85.31 $\%$ & 94.34 $\%$\\
ResNet50 & 84.55 $\%$ & 87.60 $\%$& 84.33 $\%$ & 85.21 $\%$ & \textbf{96.98} $\%$\\
VGG16 & 78.18 $\%$ & 82.04 $\%$ & 77.81 $\%$ & 78.31 $\%$ & 90.00 $\%$\\
VGG19 & 82.61 $\%$ & 83.16 $\%$ & 83.41 $\%$ & 83.19 $\%$ & 95.54 $\%$\\
MobilenetV2 & 85.00 $\%$ & \textbf{88.26} $\%$ & 84.67 $\%$ & \textbf{85.42} $\%$ & 96.61 $\%$\\
Densenet121 & 84.09 $\%$ & 87.51 $\%$ & 83.78 $\%$ & 83.91 $\%$ & 94.24 $\%$\\
\hline
\end{tabular}
\caption{Classification scores across all GCN classification models in datasets from Hospital B.}
\label{table5:Accuracies across baoan datasets of GCN}
\end{table*}

\subsection{Qualitative assessment}
\label{sec5}
Fig.~\ref{fig3:Medical concept reasoning explanation} presents interpretebiliies, i.e. concepts importances maps for mobilenetV2. from three studied graph explainers. We observe that for FASP and FTSP, three explainers generate almost same concept importance. The difference is the contribution scores of concepts. As for FASP. the contribution maps of these three techniques differ a llittle. Interestingly, all three approaches focus on the US and SP. Only Graph Grad-CAM captures the SB.

\begin{figure}[ht]
\centering
\includegraphics[width=0.4\textwidth]{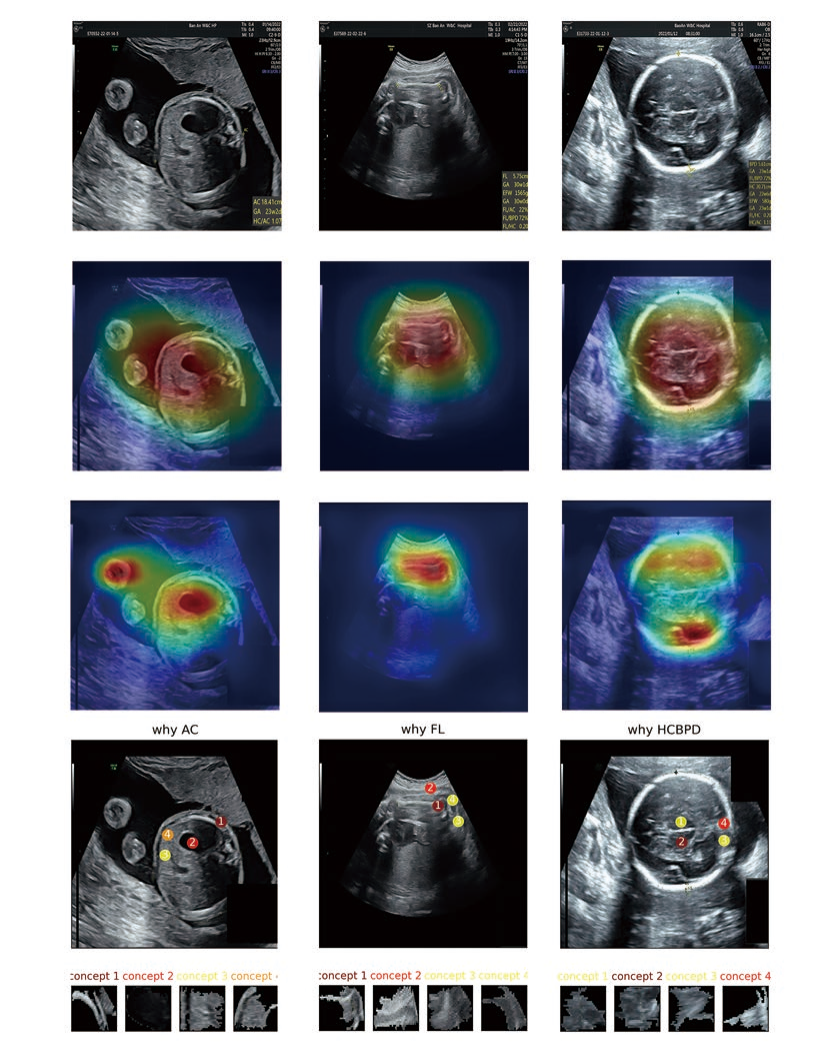} 
\caption{ Comparison of explainability methods on Fetal Standard Planes based on ResNet34.
}
\label{fig6: Comparison of ResNet34}
\end{figure}

Based on the above explanation, our algorithm can reasonably explain the logic of the decision mechanism in the network from the perspective of high-order relations and discover the reasons for the failure of the decision.

To evaluate the effectiveness of extracted medical years of clinical experience. 
We explain the classification results of MobilenetV2 and ResNet34 and compare them with other interpretation methods. We randomly selected 100 images and interpretation results from each class and showed them to doctors. For visualization of both models, participants were shown four sections: original image, interpretation results of Grad-CAM, CAMERAS, and our framework.   Fig.~\ref{fig5: Comparison of MobilenetV2} and Fig.~\ref{fig6: Comparison of ResNet34} are examples of explainability methods on Fetal Standard Planes. We asked participants to imagine facing the following situations: 

We will conduct reliability analysis for classification models (MobilenetV2 and ResNet34) trained on Fetal Standard Planes. We asked you to imagine that you are part of the team that will test this in clinical and wants to understand when the model is unreliable and perform operations upon failures. All participants were shown results from MobilenetV2 and ResNet34. They agreed that the introduction of interpretable results would enhance the trust of users in the classification model.  However, 
in contrast to many other approaches that highlight important regions, our methods construct the higher-order semantic relationships between concepts and it has obvious advantages in identifying errors. And all five doctors in the study agreed that our method was more clinically useful.

\section{Discussion}
\subsection{Clinical implication}
The localization of the standard plane is still a challenging problem, which requires the identification of complicated anatomical structures. FASP, FFSP, and FTSP are the most important views for taking measurements and assessing the fetus' health. Although current studies have achieved state-of-art-performance in the automatic localization of standard planes from US videos, they are black-box models and are difficult to trust doctors, which limits their clinical application. Our proposed approach provides an explanation of the model's decisions from the anatomical level, which greatly enhances the users' confidence. 

\subsection{Limitations}
There are several limitations to this study. First, although our method was validated in two centers, the data of Hospital B was not large, and there may be deviations. Moreover,  we just apply our approach to analyze images so far, and we haven't taken real-time medical videos into consideration.
 In our approach, the most time-consuming step is the location of anatomical structure with medical prior. Second, the high classification accuracy in our base models makes it more applicable in clinical practice, further multi-center experiments are still needed to prove the generalization performance of the method.

\section{Conclusion}
In this work, we proposed an approach to interpreting the decision mechanism of a neural network from the perspective of a sonographer's cognition. We present a medical reasoning explanation framework combining medical prior knowledge which can extract effective medical concepts for diagnosis and model the spatial relationship between them. The experiments in section~\ref{sec5}
showed that our framework can visualize the reasoning process of the neural network at the conceptual level that a sonographer can understand, which is more likely to be approved by clinicians. Furthermore, with the interpretation from the framework, we demonstrate that it can enhance the doctor's confidence in the model's prediction. The proposed interpretability method contains terms in the medical field, which is consistent with the knowledge and experience of clinicians. We believe that it can potentially be of great use in computer-aided diagnosis and contribute to the promotion and application of medical AI.


\bibliographystyle{plain} 
\bibliography{reference}

\end{document}